\documentclass[runningheads]{llncs}
\usepackage[T1]{fontenc}
\usepackage{graphicx}
\usepackage{subfigure}
\usepackage{booktabs}
\usepackage{makecell}
\usepackage[accsupp]{axessibility}  
\usepackage[pagebackref,breaklinks,colorlinks,citecolor=black]{hyperref}
\usepackage{float}
\usepackage{appendix}
\usepackage{subcaption}

\usepackage{color}

\begin{document}
\title{Guided SAM: Label-Efficient Part Segmentation}
%
%
\author{S.B. van Rooij\inst{1}\orcidID{0009-0001-0948-8175} \and G.J. Burghouts\inst{1}\orcidID{0000-0001-6265-7276}} 

\authorrunning{S.B. van Rooij and G.J. Burghouts}


\institute{TNO, The Netherlands}
\maketitle              

\begin{abstract}
Localizing object parts precisely is essential for tasks such as object recognition and robotic manipulation. 
Recent part segmentation methods require extensive training data and labor-intensive annotations. Segment-Anything Model (SAM) has demonstrated good performance on a wide range of segmentation problems, but requires (manual) positional prompts to guide it where to segment. Furthermore, since it has been trained on full objects instead of object parts, it is prone to over-segmentation of parts. To address this, we propose a novel approach that guides SAM towards the relevant object parts. Our method learns positional prompts from coarse patch annotations that are easier and cheaper to acquire. We train classifiers on image patches to identify part classes and aggregate patches into regions of interest (ROIs) with positional prompts. SAM is conditioned on these ROIs and prompts. This approach, termed `Guided SAM', enhances efficiency and reduces manual effort, allowing effective part segmentation with minimal labeled data. We demonstrate the efficacy of Guided SAM on a dataset of car parts, improving the average IoU on state of the art models from 0.37 to 0.49 with annotations that are on average five times more efficient to acquire.


\end{abstract}

\section{Introduction}
\label{sec:intro}

Precise localization of object parts is essential for many tasks, including scene perception \cite{palmer1999vision}, recognizing objects by the their parts \cite{jain1988evidence,reddy2018carfusion}, part-whole understanding \cite{biederman1987recognition,jia2020fashionpedia} and robotic manipulation \cite{myers2015affordance}. A specific part indicates what the object can do, e.g. the sharp blade of the knife can be used to cut, whereas the handle can be used to hold it. Segmentation is helpful to localize where the part is exactly, which is a requirement for a robot to grasp it at the right point, use it in the right way, or to understand the attributes of the part such as size and shape. However, segmentation of parts is not trivial. Boundaries between parts are not always clear (e.g. the hood of a car), parts can be very small compared to the full object size (e.g. the side mirror of a car), and they can have large inter-class variations (e.g. cars have very different lights). 

\begin{figure}[t]
    \centering
    \subfigure[Image and ground truth]{
        \includegraphics[width=0.33\linewidth]{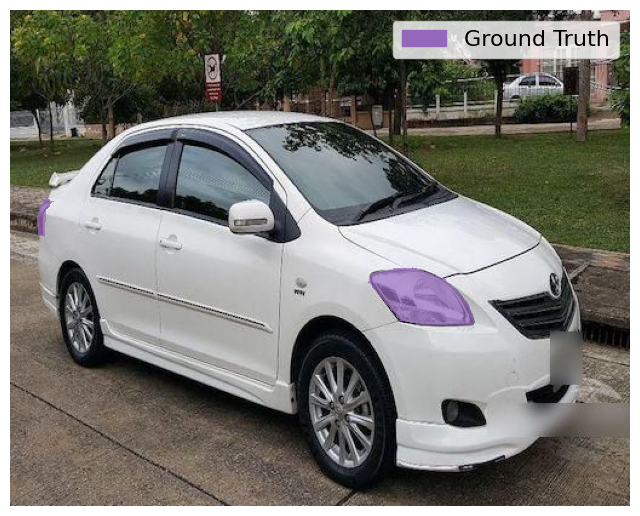}
    }
    \subfigure[Grounded SAM \cite{ren2024grounded}]{
        \includegraphics[width=0.33\linewidth]{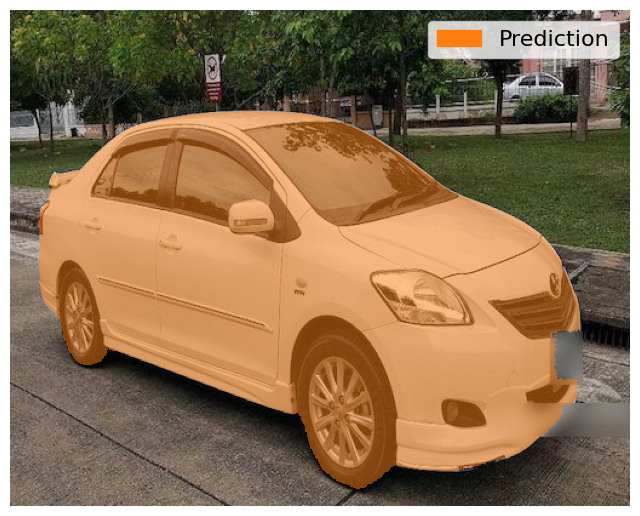}
    }
    \subfigure[VLPart \cite{sun2023going}]{
        \includegraphics[width=0.33\linewidth]{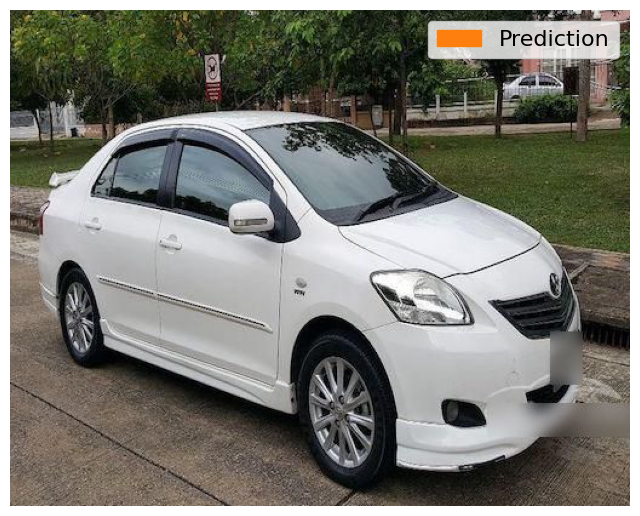}
    }\\
    \subfigure[Initial guidance]{
        \includegraphics[width=0.33\linewidth]{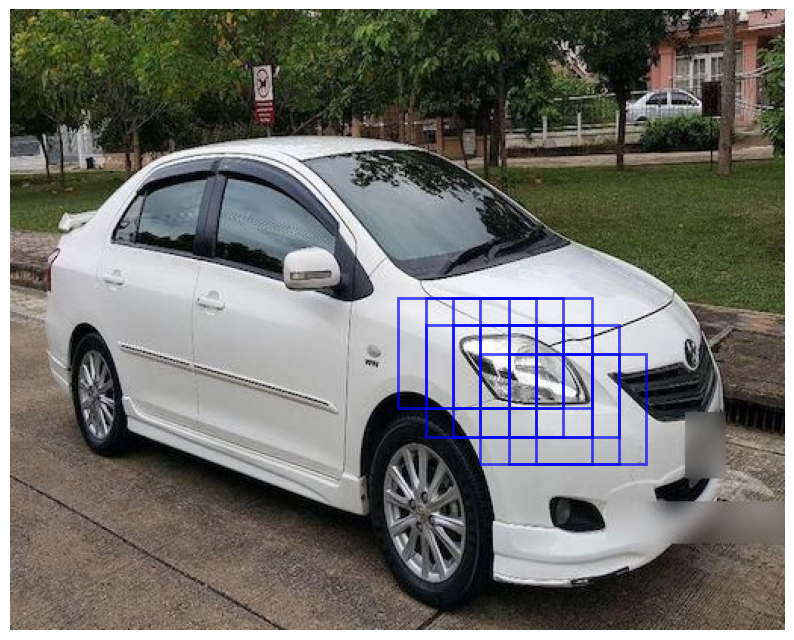}
    }
    \subfigure[Refined guidance]{
        \includegraphics[width=0.33\linewidth]{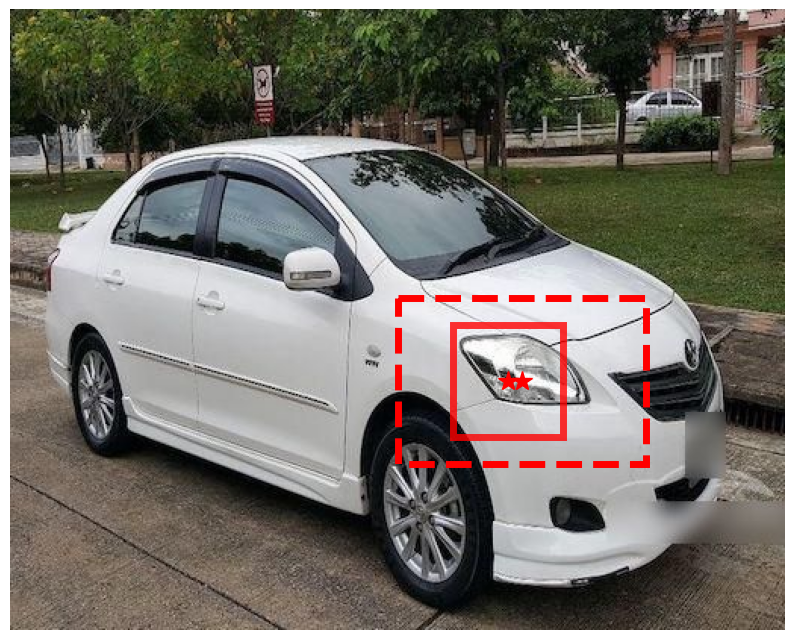}
    }
    \subfigure[Our prediction]{
        \includegraphics[width=0.33\linewidth]{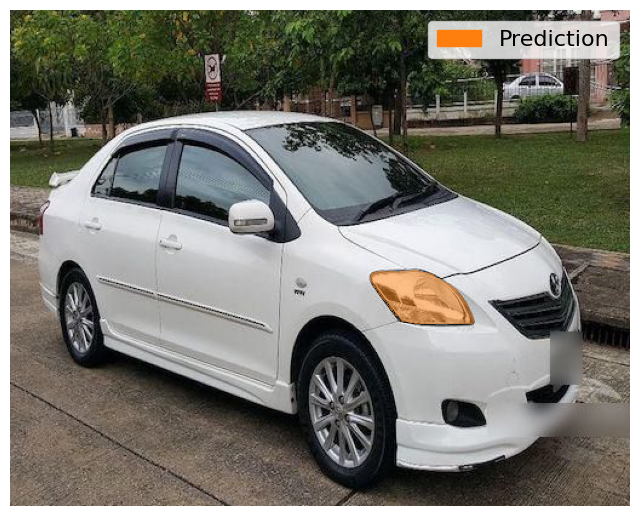}
    }
    \caption{Guiding SAM (bottom row) for part segmentation, where SOTA methods fail (top row), our patch guidance (d) and refinement (e) is more effective.}\vspace{-0.7cm}
    \label{fig:method}
\end{figure}

Recently, advanced methods have become available for part segmentation. VLPart \cite{sun2023going} trains a model on various granularities at the same time: parts, objects, and image annotations jointly provide multiscale learning signals. An object is parsed to find its parts, which provides the part segmentation with helpful contextual cues. OV-PARTS \cite{wei2024ov} builds on CLIP \cite{radford2021learning} and adapts it for part segmentation. The context of the part is provided by an object mask prompt and a compositional prompt shifts the model's attention to the parts \cite{ren2024grounded}. Grounded SAM leverages Grounding DINO \cite{liu2023grounding} as an open-vocabulary model to localize objects or parts, which are subsequently segmented by the Segment-Anything Model (SAM) \cite{kirillov2023segment}. The performance of these models is impressive. However, on common object parts they may still fail, see e.g. Figure \ref{fig:method} (b) and (c). 

Today's part segmentation models can be finetuned or retrained, but this typically requires large datasets. OV-PARTS was trained using ADE20K-Part-234 \cite{wei2024ov} and VLPart was trained using PACO \cite{ramanathan2023paco} with 641K part masks. Moreover, part masks are labour intensive annotations, i.e. pixel-precise masks. Therefore, improving the models on specific parts of interest involves large datasets or labour intensive labelling. Our objective is a methodology that requires low amounts of labelled images, and moreover, annotations that are easy to acquire with a few clicks per image.

Our starting point is the Segment-Anything Model (SAM) \cite{kirillov2023segment}, because it has demonstrated a very strong performance on a wide range of image contents and across various granularities from objects to parts. But, SAM is not directly applicable to part segmentation, because it requires manual guidance where to segment. This guidance comes in the form of one or more locations in the image, which are referred to as positional prompts. We want to substitute the manual prompting by automated prompting, such that the part segmentation can be performed in a fully automated manner. This positional prompting is tailored to the part of interest. Our approach is to \textit{learn} the positional prompting, from coarser annotations that are easy to acquire. Coarse annotations come from image patches of approximately 1/14th of the image width and height. The annotation says whether a patch contains the part. Such patch annotations are much coarser and simpler to acquire than pixel-precise masks, therefore this strategy is significantly more efficient. To advance the efficiency further, we leverage prototypical patches \cite{nauta2023pip} that group the parts already reasonably well before annotation. For each part of interest, we learn a patch classifier to predict whether a test patch contains the part. For the representation of a patch, we use DINOv2 for its strong representational power for a wide variety of image contents \cite{oquab2023dinov2}. For a sense of context, the predicted patches are locally grouped into regions of interest (ROIs). For the positional prompt within the ROI, a location is inferred using a maximum likelihood formulation. SAM is invoked on the ROI with the positional prompt. The advantage of a ROI is two-fold: it provides a contextual cue and avoids the necessity to process the full image. Processing only the ROI is advantageous for reducing computations and avoiding false alarms in irrelevant image regions. We coin our method `Guided SAM' and it is illustrated in Figure \ref{fig:method}.

The efficacy of Guided SAM is measured on a dataset of car parts. This is an interesting testset, because the parts vary significantly in size, from very small (a tiny back light), small (side mirror, front light), medium (bumper, trunk) to large (door, hood). We compare our method with recent models that have shown impressive performance, namely vision-language models that take textual prompts: Grounded SAM \cite{ren2024grounded} and VLPart \cite{sun2023going}. Also, we compare various positional prompting strategies combined with SAM \cite{kirillov2023segment}. We will show the efficiency of acquiring patch annotations and their suitability for Guided SAM. It is possible to learn a good segmentation model for a part from only 16 to 64 images, which outperforms state of the art (SOTA) models, while requiring only 5 clicks per image on average.


\section{Related Work}
\label{sec:rel_work}

For part segmentation, vision-language models have been proposed recently, which can be prompted with a textual description of the part. VLPart \cite{sun2023going} trains the model on the part-, object- and image-level to align language and image. An object is parsed by dense semantic correspondence. This approach benefits from various data sources and foundation models, as demonstrated in experiments where the model was applied to unseen object-part combinations (open-vocabulary). OV-PARTS \cite{wei2024ov} modifies and tailors CLIP \cite{radford2021learning} for part segmentation. An object mask prompt is proposed to enable the model to take the context into account. To attend more to the parts than whole objects or scenes, a compositional prompt was proposed to reshift attention \cite{ren2024grounded}. Since OV-PARTS and VLPart are both designed to perform open-vocabulary part segmentation, we only consider VLPart in our experiments. 

Grounded SAM \cite{ren2024grounded} combines two powerful models: Grounding DINO \cite{liu2023grounding} and SAM \cite{kirillov2023segment}. Grounding DINO localizes boxes in the image based on a textual description. Each box contains a prediction where the target may be, in the form of a initial mask. This box is represented by an embedding pair of the top-left corner and the bottom-right corner that serve as positional prompts. These prompts are provided to SAM \cite{kirillov2023segment}, which segments the target. 

Grounded SAM \cite{ren2024grounded} inspired us to look more deeply into the positional prompts themselves. Rather than using the box representations as input for SAM, we aim for a regional prediction that is centered around the part already, to acquire a small but tailored sense of context. Moreover, we want to predict the positional prompts more precisely. For that purpose, we take inspiration from OV-PARTS \cite{wei2024ov} and Grounded SAM \cite{ren2024grounded}, by following their strategy to incorporate some image context around the part. Instead of an implicit context via multiscale annotation (VLPart \cite{sun2023going}), we follow OV-PARTS and Grounded SAM by providing an explicit context in the form of a mask or box. Our ROI approach differs because the ROI is more centered around the part, instead of the full object.

\section{Guided SAM}
\label{sec:method}

Our method segments parts of objects, such as the light of a car, see Figure \ref{fig:method} (a). Two state of the art methods, Grounding SAM and VLPart, fail on this task, leading to respectively false positives in Figure \ref{fig:method} (b) and false negatives in Figure \ref{fig:method} (c). Our objective is to train a capable part segmentation model, while requiring a small amount and labour-efficient type of human annotations. For label-efficiency, we leverage a model $M$ that has strong performance on segmentation already: SAM \cite{kirillov2023segment}. This model cannot be applied directly to an image in order to segment a specific part. It requires a spatial cue, provided as a positional prompt $P_{(x,\,y)}$ (a pixel location). Our rationale is to \textit{learn} the spatial cue for a part, in order to guide SAM towards regions in the image where the part is located, $P_{ROI}$. Our guidance model $\mathcal{G}$ takes an image $I$ and a part $C$ and produces a set of tuples: 

\begin{equation}
    \mathcal{G}(I\,|\,C) \rightarrow \{(P_{ROI}^i,\,P_{(x,\,y)}^i)\}_{i \in 1:N}
\end{equation}

Here, $P_{ROI}^i$ serves as a region of interest (ROI) that conditions where SAM is applied. For each $P_{ROI}^i$, $P_{(x,\,y)}^i$ serves as the positional prompt for SAM to segment the part. The guidance model $\mathcal{G}$ involves a learner $\mathcal{L}$ that classifies whether an image patch $p^j$ contains the part $C$: $\mathcal{L}(I\,|\,p^j) \rightarrow c$, where $c$ is the confidence for the part class. Classifying patches is a much simpler learning task than predicting pixel-precise segments. Moreover, the learning requires a simpler form of annotation, i.e., a binary label for the patch if it contains the part or not. Our hypothesis is that such a patch classifier can be learned with a small amount of labels that are simple to annotate. Figure \ref{fig:method} (d) shows the classified patches that are likely to contain the part. A ROI $P_{ROI}^i$ is generated by grouping the classified patches. The positional prompt $P_{(x,\,y)}^i$ for each ROI $P_{ROI}^i$ is inferred from its constituent patches and their respective confidences. Figure \ref{fig:method} (e) shows $(P_{ROI}^i, P_{(x,\,y)}^i)$ that was inferred from the classified patches in Figure \ref{fig:method} (d).

For the segmentation of an object part, both the ROI $P_{ROI}^i$ and the positional prompt $P_{(x,\,y)}^i$ are used. SAM is conditioned on $P_{ROI}^i$, by passing only the respective image contents. This avoids false positives at irrelevant image regions. SAM is also conditioned on $P_{(x,\,y)}^i$, in order to give it a good starting point for segmentation. Figure \ref{fig:method} (f) shows the part segmentation. Our method enables to use SAM for part segmentation after providing a few labeled patches. 

\begin{figure}[h!]
    \centering
    \subfigure[Annotation with prototypical patches as guidance.]{\includegraphics[width=0.75\textwidth]{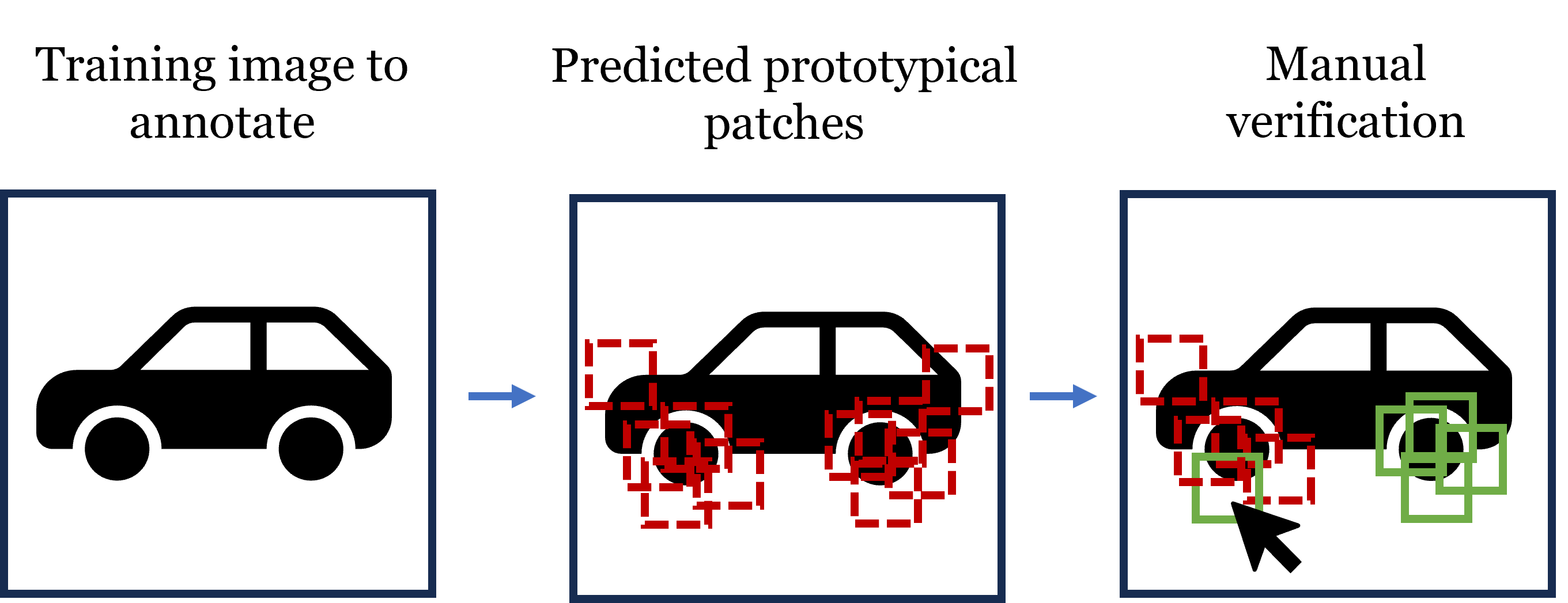}}
    \label{fig:method_annotation}
        
    \vspace{1em} 
    
    \subfigure[Inference with GuidedSAM before segmentation.]{\includegraphics[width=\textwidth]{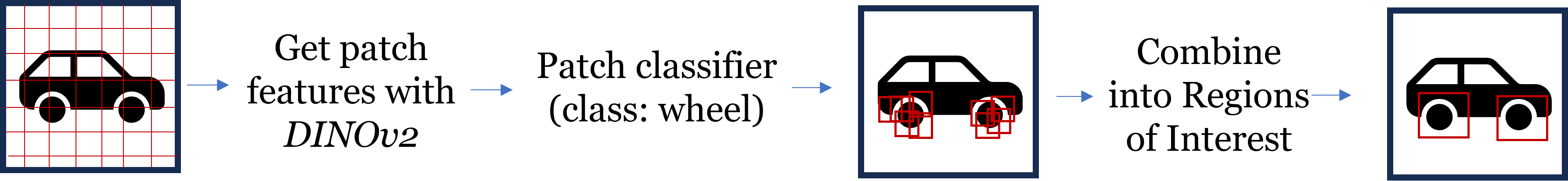}}
    \label{fig:method_inference}
    
    \vspace{1em} 
    
   \subfigure[Segmentation using different model variants.]{\includegraphics[width=0.7\textwidth]{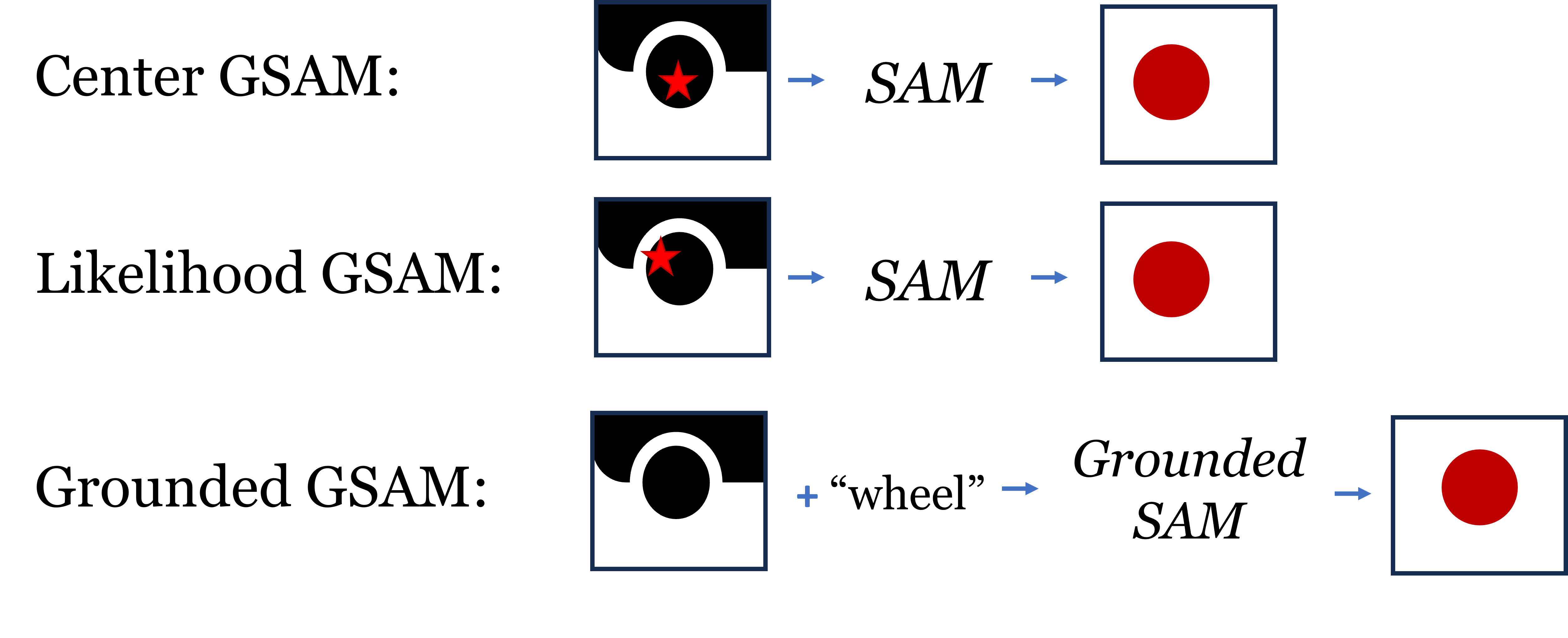}}
    \label{fig:method_models}
    
    \caption{Various elements of the pipeline for GuidedSAM, showing the efficient annotation process in (a), the inference until the segmentation in (b) and the segmentation of the regions of interest with different model variants in (c).}
    \label{fig:pipeline}
\end{figure}

The flow diagram of GuidedSAM is illustrated in Figure \ref{fig:pipeline}. The annotation of patches is shown in Figure \ref{fig:pipeline} (a) and will be further explained in Section \ref{sec:prototypes}. The inference steps before the segmentation are shown in Figure \ref{fig:pipeline} (b) and will be covered in Section \ref{sec:classifier}. Finally, Figure \ref{fig:pipeline} (c) shows the guided segmentation with multiple model variants that will be explained in Sections \ref{sec:GSAM} and \ref{sec:model_variants}.

\subsection{Prototypical Patches}
\label{sec:prototypes}
\begin{figure}[h!]
    \centering
    \includegraphics[width=0.90\textwidth]{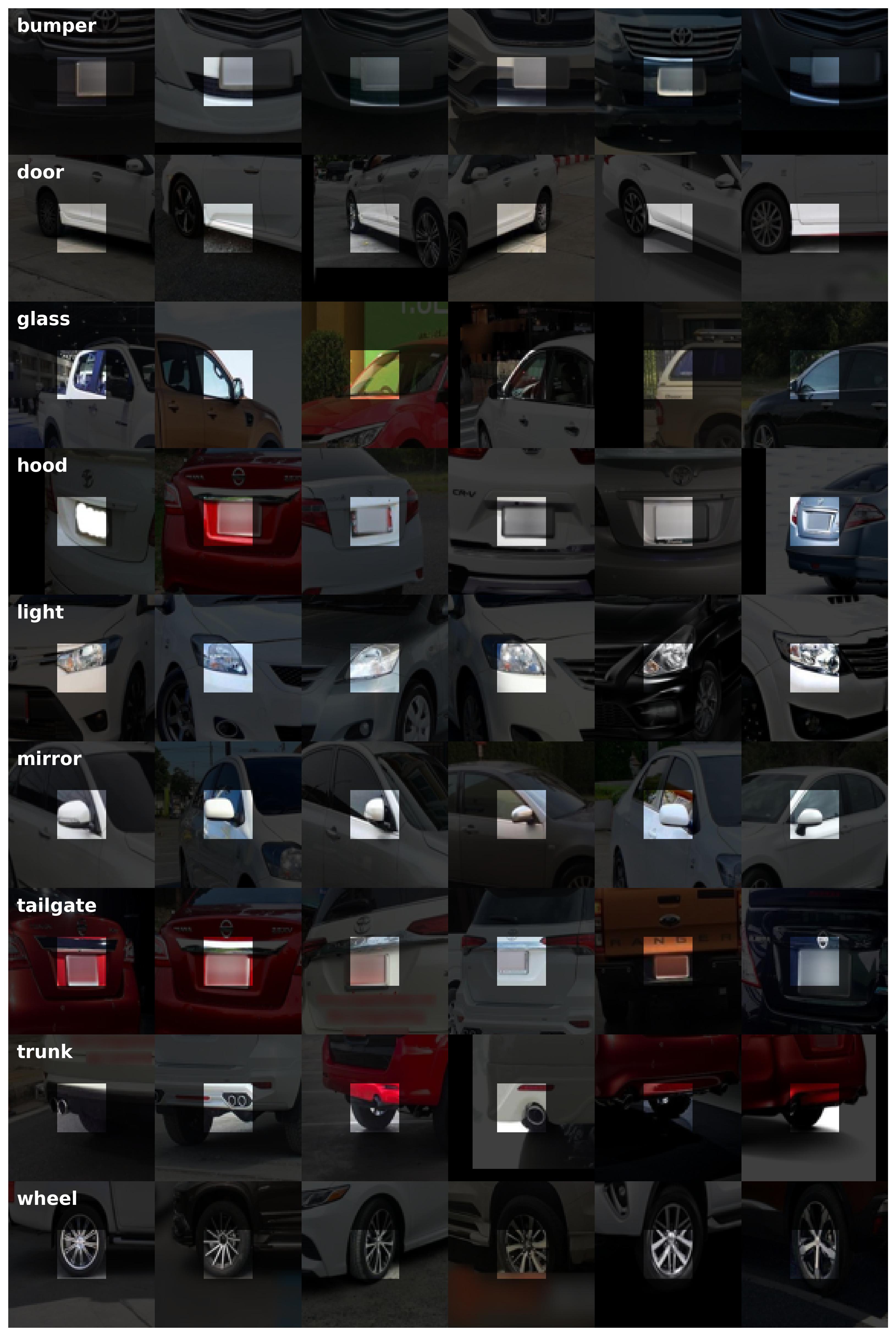}
    \caption{Prototypical patches group together similar object parts, which facilitates the annotation.}
    \label{fig:prototypes}
\end{figure}

Our learner $\mathcal{L}$ requires a set of binary labels for respective patches whether they contain the part of interest: $\mathcal{D} = \{(z_i, l_i)\}_{i=1}^{M}$ with $M$ samples, each consisting of a patch $z_i$ and a label $l_i \in [0, 1]$ indicating presence of the part. To arrive at $\mathcal{D}$, the problem is that patches containing parts have a low prevalence, considering that the parts are typically small. Drawing a random selection of patches for annotation, is not efficient. Instead, we select patches that have a larger probability of containing the part. We group similar patches by means of prototypical patches \cite{nauta2023pip}. The prototypes do not have a name, neither are they necessarily related to the part of interest. To relate the prototypes to the part, we match each prototype to the part name, using the visual-textual similarity measure of CLIP \cite{radford2021learning}. Each prototype is assigned a score for the part of interest. Figure \ref{fig:prototypes} shows examples for various car parts, illustrating that the prototypes group together patches that relate to the respective parts. 

For a specific part, the prototypes are ranked by descending CLIP score. Each prototype is verified by a human annotator. For illustration purposes, we indicate this for an example image for the part `wheel' in Figure \ref{fig:pipeline} (a). This involves one affirmative click if all patches of the prototype contain the part. Similarly, one negative click is required when none of the patches contain the part. More clicks are needed when most patches contain the part, by negating the fewer patches that do not contain it, or vice versa. This procedure yields $(z_i, l_i)$ that constitute $\mathcal{D}$.

\subsection{Guidance Classifier}
\label{sec:classifier}

Given $\mathcal{D} = \{(z_i, l_i)\}_{i=1}^{M}$, the classifier $\mathcal{L}$ is learned, which predicts for a test patch $z_j$ the probability that it contains the part. The patch $z_j$ is represented as a feature vector by a model $\phi(\cdot)$: $z_j^\phi = \phi(z_j)$. For $\phi(\cdot)$ we consider DINOv2 \cite{oquab2023dinov2} which has proven to be a robust feature extractor. $\mathcal{L}_p$ is an SVM \cite{cortes1995support} with a radial basis function as the kernel. The parameters are learned from train samples $\{(z_i^\phi, l_i)\}$. During inference, the trained classifiers are used to predict a rough location of the parts of interest in the test image. This process is illustrated on the left side of Figure \ref{fig:pipeline} (b). 

\subsection{Guided Segmentation}
\label{sec:GSAM}
A ROI $P_{ROI}^i$ is generated by grouping the predicted patches $\{p_j\}$ that are likely to contain the part: $\{\mathcal{L}(I\,|\,p_j) > c_t\}$, where $c_t$ is a threshold on the confidence $c$. An example is provided in Figure \ref{fig:method} (d). The grouping is based on the patches from $\{p_j\}$ that overlap: $\{p_k\}_{IoU\,>\,0}$, where $\{p_k\} \subset \{p_j\}$. The ROI $P_{ROI}^i$ is the combination of the minimum and maximum coordinates of the patches in $\{p_k\}$. This is also shown in Figure \ref{fig:pipeline} (b), where the individual patches are combined into larger ROIs that take into account more context. The positional prompt $P_{(x,\,y)}^i$ for each ROI $P_{ROI}^i$ is inferred from its constituent patches $p_k$ and their respective confidences $c_k$ by taking the center coordinates of the patch with the highest confidence. 
Figure \ref{fig:method} (e) shows $(P_{ROI}^i, P_{(x,\,y)}^i)$. Guided SAM is the conditioning of SAM on $(P_{ROI}^i, P_{(x,\,y)}^i)$. An example result is shown in Figure \ref{fig:method} (f). 

\subsection{Model Variants}
\label{sec:model_variants}

Besides the version of Guided SAM described in Section \ref{sec:GSAM}, we also consider other variants of our conditioning. We make a distinction between applying the segmentation on the ROI $P_{ROI}^i$ (i.e. the combination of the classified patches) or taking the individual patches $p_k$ as ROIs. This is also depicted in Figure \ref{fig:method} (e), where the bounding box with the dashed line stands for a ROI of combined patches and the smaller box represents an individual patch. For these ROI types there are various options to segment or prompt. Firstly we can replace the segmentation model SAM by Grounded SAM \cite{ren2024grounded} and apply it to the ROI: we coin this model Grounded Guided SAM (GGSAM). This version takes a textual prompt instead of the positional prompt $P_{(x,\,y)}^i$. Secondly, we can infer the positional prompt $P_{(x,\,y)}^i$  from the center coordinates of the ROI, which we coin Center Guided SAM (CGSAM). The version that was described before, where the positional prompt is inferred from the center of the patch with the maximum confidence in $P_{ROI}^i$, is coined Likelihood Guided SAM (LGSAM). This method can only be applied to $P_{ROI}^i$, since the other ROI is just a single patch. These model variants are illustrated in Figure \ref{fig:pipeline} (c) on the combined ROIs.

\subsection{Computational load}
The computational steps for model inference are shown in Figure \ref{fig:pipeline} (b). These computations are required on top of the original SAM. We apply an efficient DINOv2 \cite{oquab2023dinov2} variant to compute the patch features, i.e. ViT-B, which has only 86M parameters. For each part class, the same DINOv2 features are re-used, with a class-specific part classifier. This classifier is an SVM, which involves negligible computations compared to SAM. SAM has 94.7M parameters, comparable to DINOv2 ViT-B, so the computation time of our Guided SAM will be approximately doubled by the classifier guidance. If computational efficiency is essential, faster alternatives are available, e.g. \cite{zhao2023fast}, which has a faster backbone. Currently our method applies SAM to every region of interest that is proposed by the part classifiers. This can be implemented more efficiently by re-using its feature maps and only re-running SAM's efficient head on the various regions of interest.

\section{Experiments}
\label{sec:experiments}

\subsection{Setup}
For evaluation we consider the Car Parts Segmentation dataset \cite{DSMLR_Carparts}, because of its large inter-class and intra-class variations. The part classes have very different sizes relative to the object. The same part can have different appearances, e.g. forms, sizes and colors. The dataset contains 400 images with annotated segmentation masks of 18 part classes. We merged the different sides (front vs. back, left vs. right) to one part class. There are a total of 9 part classes: bumper, glass, door, light, hood, mirror, tailgate, trunk, and wheel. For language-guided methods (i.e. VLPart and Grounded SAM) we made slight variations to these class names that better reflect the nature of the classes (e.g. replacing glass for window). As a metric, we consider the IoU for each part. The training efficiency is established by increasing the number of training images from 1, 2, 4, 8, ..., 64.

\subsection{Patch Selection}
\begin{figure}[!h]
    \centering
    \subfigure[Retrieval efficacy.]{
        \includegraphics[width=0.48\textwidth]{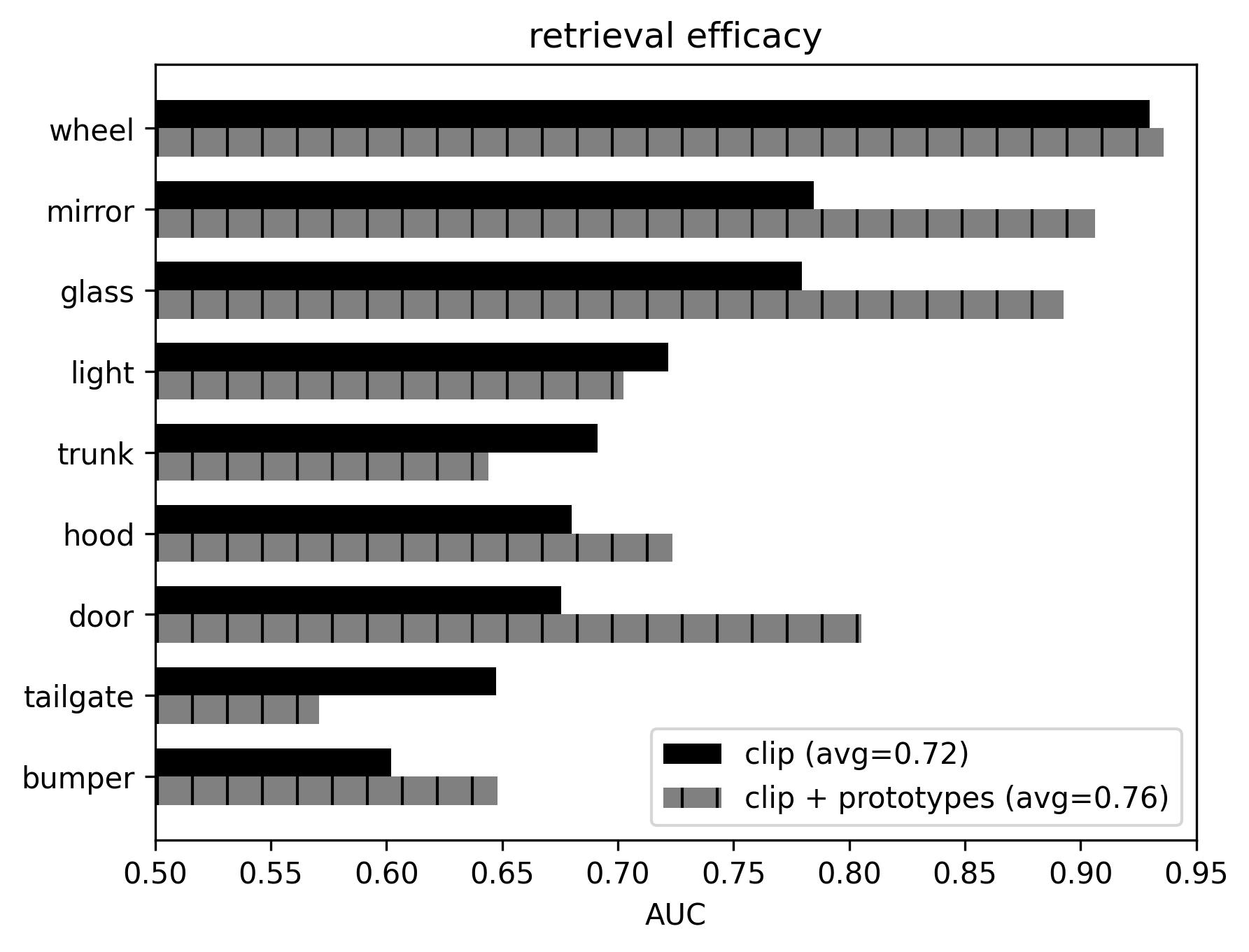} 
    \label{fig:retrieval_efficacy}} 
    \subfigure[Annotation efficiency.]{
        \includegraphics[width=0.472\textwidth]{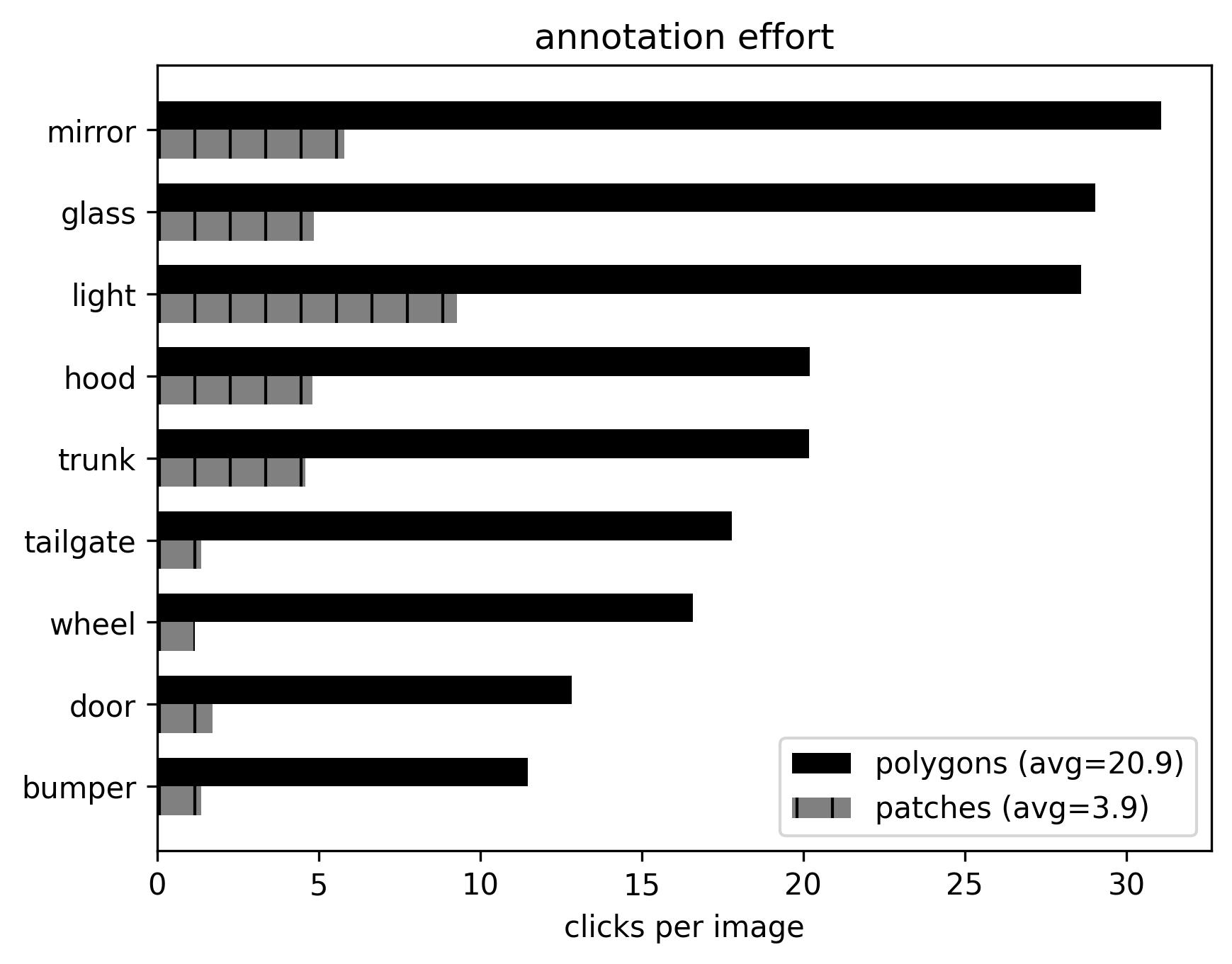}
    \label{fig:annotation_effort}}
    \caption{Prototypical patches are helpful to find the parts (a). Annotating patches is on average >5x more efficient (b). }
\end{figure}

For finding the patches that contain the part, we evaluate the merit of the prototypical patches. To that end, we compare the retrieval efficacy of CLIP with and without the prototypes. Figure \ref{fig:retrieval_efficacy} shows that for most part classes, there is an advantage to consider the prototypical patches. This means that it is helpful to consider the average CLIP score for each prototype before ranking. 

To evaluate the annotation efficiency, we compare the amount of manual clicks that are necessary for conventional annotation of polygons to create pixel masks, and for our patch-based annotations using the prototypes. Figure \ref{fig:annotation_effort} shows the annotation efforts for both strategies for the various part classes. On average, annotating patches with our prototype strategy is more than 5 times more efficient. The speedup is most prominent for tailgate and wheel. For conventional annotation, bumper involves the least clicks, because it has a simple shape. Even compared to bumper, all parts are  annotated with fewer clicks per image when using the patch-based strategy.

\subsection{Guidance Classifier}
As shown in Table \ref{tab:guidance_accuracy}, the learned patch classifier $\mathcal{L}_p$ performs classification of object parts very accurately. For `door' the performance is the highest: AUC = 0.994 when using 64 training images (in following subsections we will experiment with fewer training images). `Wheel' also has a very high performance: AUC = 0.990, possibly because of its distinct visual features. For `trunk' the performance is the lowest, but still very high: AUC = 0.977. Trunk does not have a distinctive boundary and the part is often a flat surface without much texture or distinctive visual features. Light is also somewhat harder to classify (AUC = 0.979). It is a very small part and shows a lot of intra-class variation, such as different shape, size, color (depending on whether it is on or off).

\begin{table}[H]
\centering
\caption{The patch classifier performs very accurate classification of object parts (average AUC$\approx$0.985).}\vspace{0.15cm}
\setlength{\tabcolsep}{4.5pt} 
\begin{tabular}{@{}l@{\hspace{20pt}}ccccccccc@{}}
\toprule
\textbf{Class} & door & wheel & mirror & hood & glass & tailgate & bumper & light & trunk \\ \midrule
\textbf{AUC} & 0.994 & 0.990 & 0.989 & 0.988 & 0.988 & 0.987 & 0.986 & 0.979 & 0.977 \\ \bottomrule
\end{tabular}
\label{tab:guidance_accuracy}
\end{table}



\subsection{Comparison to SOTA}
We compare against two methods: VLPart\footnote{For VLPart we use a confidence threshold of 0.5. For the results of VLPart with varying confidence thresholds, see Supplementary Material.} \cite{sun2023going} and Grounded SAM \cite{ren2024grounded}. Our method is trained on 64 images. In following subsections we evaluate the impact of having fewer training images. Table \ref{tab:sota} reveals that Guided SAM outperforms VLPart and Grounded SAM for most object parts. Overall it is the best performer, on average IoU = 0.493 compared to 0.370 (VLPart) and 0.124 (Grounded SAM). VLPart does perform best at larger or common parts, such as wheel, door and mirror. It is surprising that VLPart does not perform better at other parts, given that it was trained on datasets that include all parts from Table \ref{tab:sota}, i.e. LVIS \cite{gupta2019lvis} and PACO \cite{ramanathan2023paco}. Grounded SAM performs much worse across the board, probably because it is optimized for objects and not for object parts, although on some parts it performs somewhat better: e.g. wheel and bumper. These are larger parts or parts with clear boundaries. Some parts have a very low performance for both VLPart and Grounded SAM: light, tailgate, and trunk, AUC$\approx$0.04. For these parts, Guided SAM performs much better: AUC$\approx$0.35. 

\begin{table}[]
\centering
\caption{Performance of VLPart \cite{sun2023going}, Grounded SAM \cite{ren2024grounded}, and Guided SAM on the Car Parts dataset in terms of IoU. Bold numbers indicate the best performance per part for the three methods. Guided SAM outperforms VLPart and Grounded SAM for most object parts.}\vspace{0.15cm}
\begin{tabular}{@{}lccc@{}}
\toprule
 & \makecell{VLPart} & \makecell{Grounded SAM} & \makecell{Guided SAM (ours)} \\ \midrule
wheel & \textbf{0.800} & 0.305 & 0.683 \\
glass & 0.621 & 0.089 & \textbf{0.638} \\
door & \textbf{0.736} & 0.202 & 0.635 \\
bumper & 0.027 & 0.299 & \textbf{0.605} \\
hood & 0.440 & 0.089 & \textbf{0.553} \\
light & 0.000 & 0.041 & \textbf{0.377} \\
tailgate & 0.000 & 0.035 & \textbf{0.370} \\
trunk & 0.006 & 0.048 & \textbf{0.314} \\
mirror & \textbf{0.696} & 0.009 & 0.259 \\ \midrule
average & 0.370 & 0.124 & \textbf{0.493} \\ \bottomrule
\end{tabular}
\label{tab:sota}
\end{table}

\begin{figure}
    \centering
    \parbox{0.99\linewidth}{
        \centering
        \includegraphics[width=0.32\linewidth]{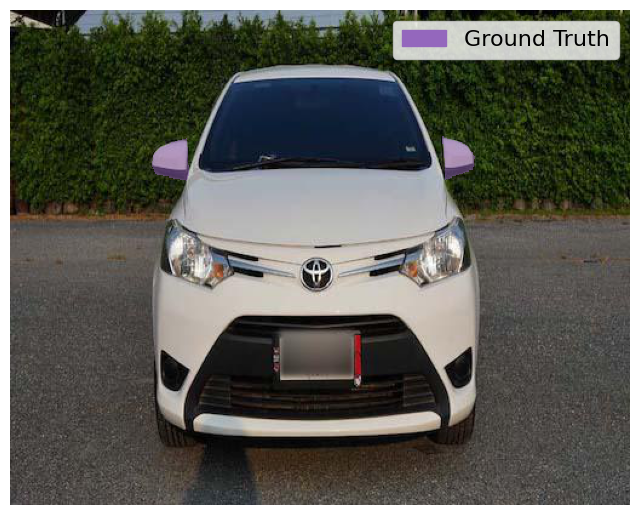}
        \includegraphics[width=0.32\linewidth]{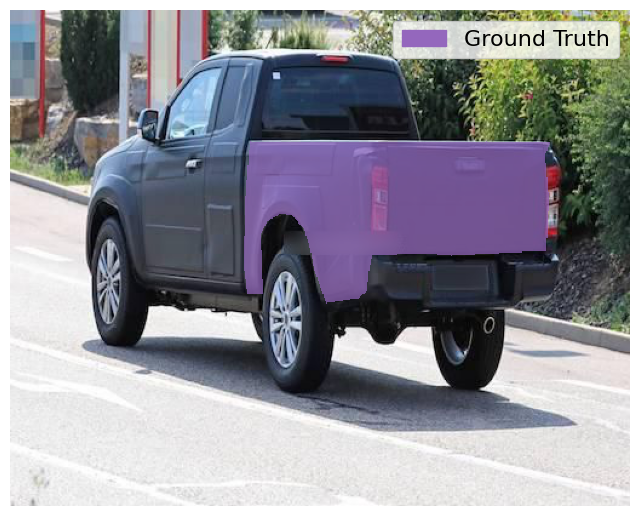}
        \includegraphics[width=0.32\linewidth]{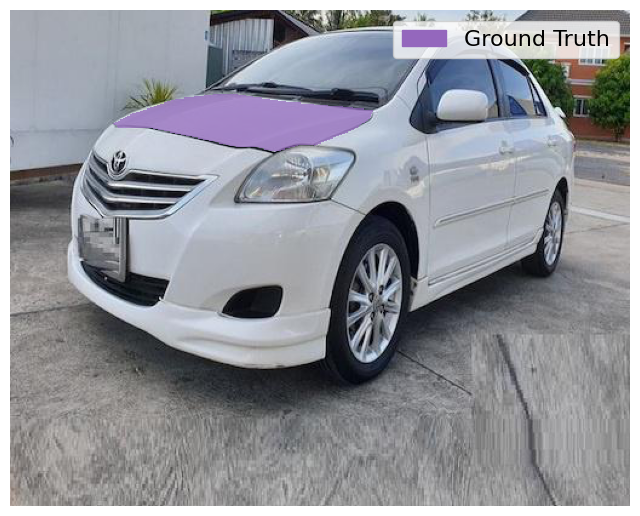}
        (a) Ground Truth \\ 
    }\\
    \parbox{0.99\linewidth}{
        \centering
        \includegraphics[width=0.32\linewidth]{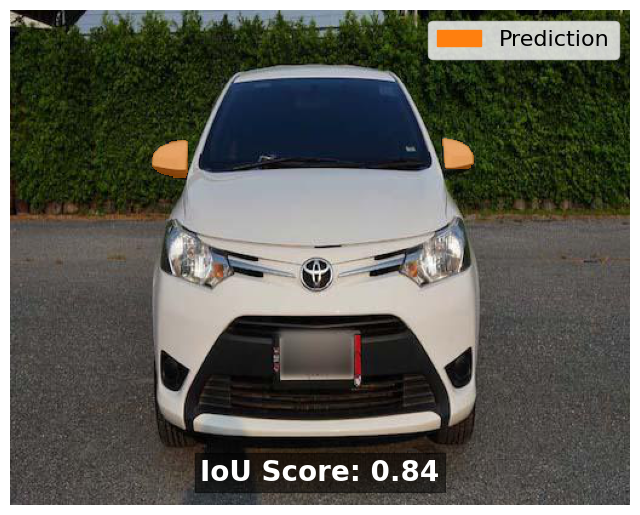}
        \includegraphics[width=0.32\linewidth]{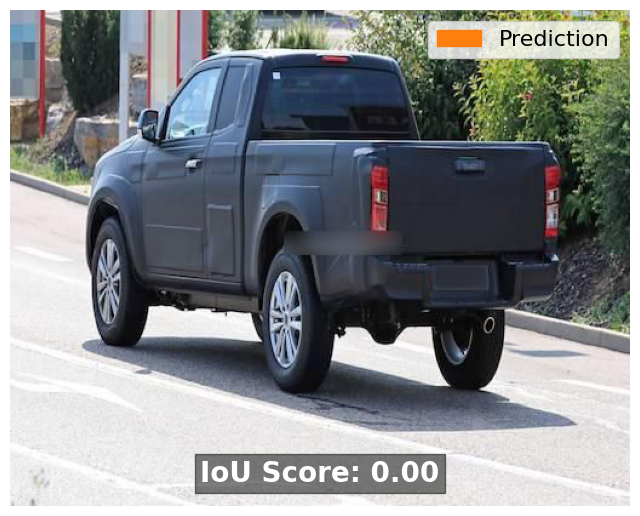}
        \includegraphics[width=0.32\linewidth]{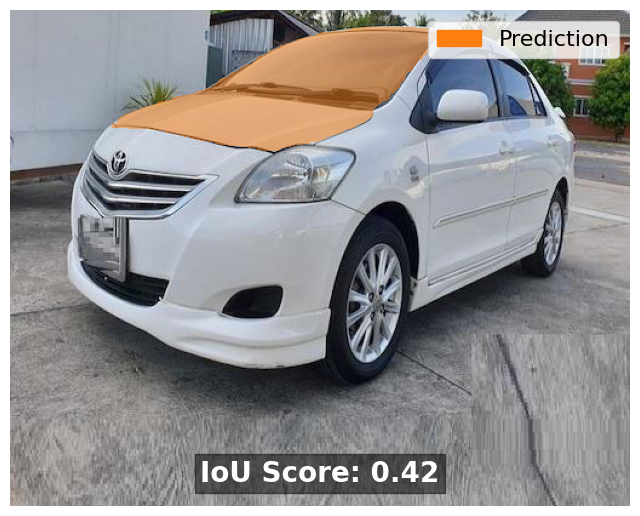}
        (b) VLPart \cite{sun2023going} \\
    }\\
    \parbox{0.99\linewidth}{
        \centering
        \includegraphics[width=0.32\linewidth]{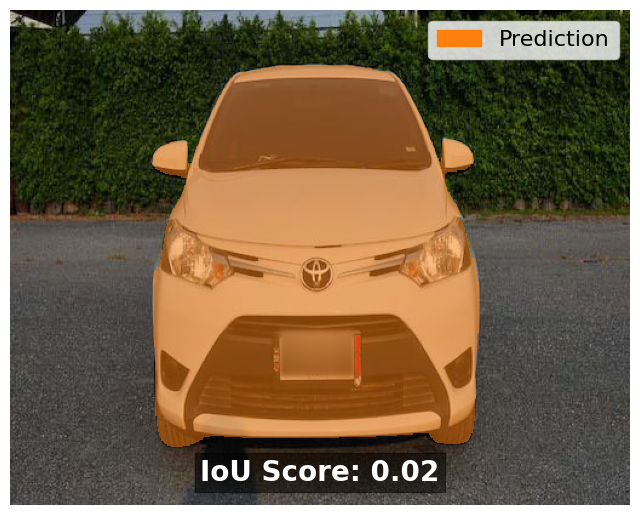}
        \includegraphics[width=0.32\linewidth]{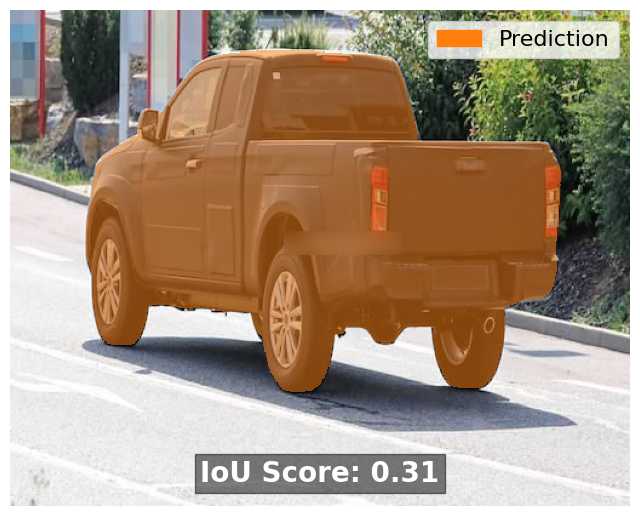}
        \includegraphics[width=0.32\linewidth]{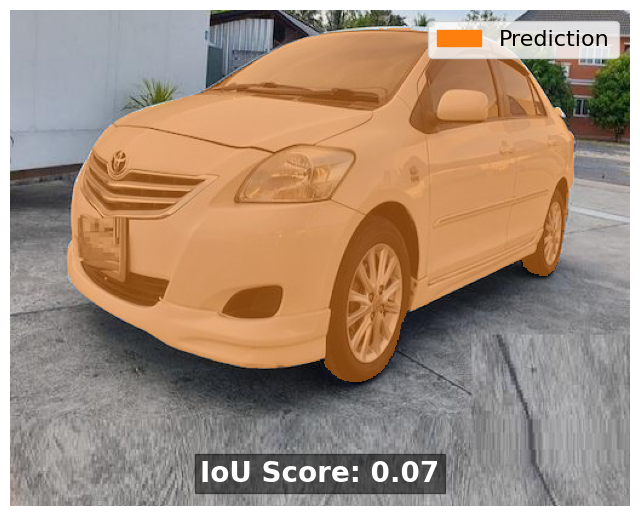}
        (c) Grounded SAM \cite{ren2024grounded} \\
    }\\
    \parbox{0.99\linewidth}{
        \centering
        \includegraphics[width=0.32\linewidth]{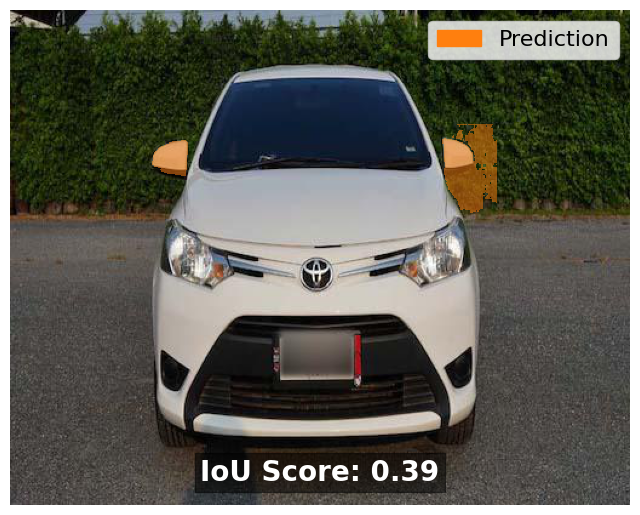}
        \includegraphics[width=0.32\linewidth]{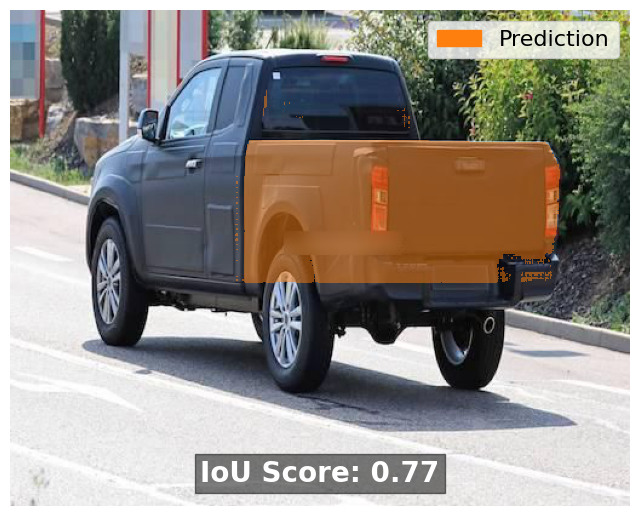}
        \includegraphics[width=0.32\linewidth]{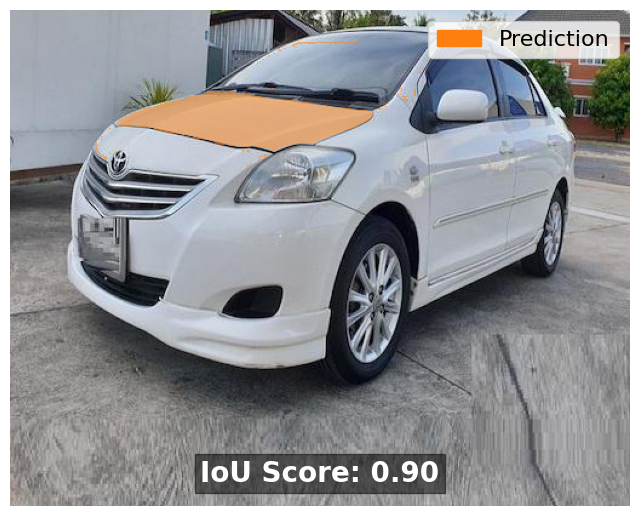}
        (d) Guided SAM (ours) \\
    }\vspace{-0.1cm}
    \caption{VLPart either segments the object part very well or misses it completely, whereas Grounded SAM typically over-segments severely and often segments the full object. Guided SAM provides a balance, often segmenting the part well, while sometimes over-segmenting.}
    \label{fig:predictions_illustrations}
\end{figure}

Predictions of the tested models are illustrated for three examples, see Figure \ref{fig:predictions_illustrations}. The top row indicates the ground truth, where the other rows show the predicted part segments. VLPart (b) is sometimes very impressive (left), while at other times it misses the part completely (middle), or over-segments it (right). Grounded SAM (c) typically segments the full objects rather than the part. Guided SAM provides a balance, often segmenting the part well, while sometimes over-segmenting or segmenting the background rather than the part.

\subsection{Evaluating Model Variants}
\label{sec:model_variants_eval}
We evaluate the model variants from Section \ref{sec:model_variants}. As a short recap, we have two main divisions: taking the $P_{ROI}^i$ as the ROI, or its consituent patches $\{p_k\}$ as individual ROIs. This is the top row in Table \ref{tab:model_variants}. For each ROI type, there are various options to segment or prompt: Grounded Guided SAM (GGSAM) which uses Grounded SAM as the segmenter, Center Guided SAM (CGSAM) which uses the ROI center as the positional prompt, and Likelihood Guided SAM (LGSAM) which uses the most likely location (i.e. the center of the patch with the highest confidence) as the positional prompt. To establish the effect of the segmentation methods, we also compare with assigning the full patch as the segment, i.e. no segmentation. We refer to this variant as Naive. 

Table \ref{tab:model_variants} presents the IoU scores per part for the model variants. ROI guidance is more effective than patch guidance, in most cases. The exceptions are tailgate and trunk, but for these parts ROI guidance performs similarly. Using a positional prompt based on the likelihood (LGSAM) is best on average. There is no single model variant that performs best for all parts. CGSAM performs best on light, hood and mirror. GGSAM appears to perform well on larger car parts that have a distinct boundary, such as bumper, door and wheel. Interestingly, the performance of GGSAM (i.e. Grounded SAM as segmenter) is much better than applying Grounded SAM on the full image, i.e. without our guidance (Table \ref{tab:sota}). We conclude that our guidance is also helpful for an existing model. 

\begin{table}[htbp!]
\centering
\caption{Performance for Region-of-interest (ROI) and Patch guidance for Grounded Guided SAM (GGSAM), Center Guided SAM (CGSAM), and Likelihood Guided SAM (LGSAM) in terms of IoU. Bold numbers indicate the best performance per part for all model variants. ROI guidance is more effective than patch guidance, where segmentation based on likelihood (LGSAM) is best on average.} \vspace{0.25cm}
{\scriptsize
\begin{tabular}{@{}lcccccc@{}}
\toprule

 & \multicolumn{3}{c}{Region-of-interest} & \multicolumn{3}{c}{Patches} \\
 \cmidrule(lr){2-4}
 \cmidrule(lr){5-7}

    & \multicolumn{1}{l}{GGSAM} & \multicolumn{1}{l}{CGSAM} & \multicolumn{1}{l}{LGSAM} & \multicolumn{1}{l}{Naive} & \multicolumn{1}{l}{GGSAM} & \multicolumn{1}{l}{CGSAM} \\ \midrule

bumper   & \textbf{0.605}            & 0.319                     & 0.423                     & 0.487                       & 0.560                                                   & 0.550                                                                        \\
glass    & 0.317                     & 0.626                     & \textbf{0.638}            & 0.354                       & 0.458                                                   & 0.480                                                                        \\
door     & \textbf{0.635}            & 0.447                     & 0.399                     & 0.440                        & 0.508                                                   & 0.582                                                                        \\
light    & 0.173                     & \textbf{0.377}            & 0.371                     & 0.170                        & 0.211                                                   & 0.217                                                                        \\
hood     & 0.395                     & \textbf{0.553}            & 0.505                     & 0.362                       & 0.406                                                   & 0.478                                                                        \\
mirror   & 0.063                     & \textbf{0.259}            & 0.206                     & 0.102                       & 0.113                                                   & 0.148                                                                        \\
tailgate & 0.325                     & 0.165                     & 0.370            & 0.318                       & 0.348                                                   & \textbf{0.389}                                                                        \\
trunk    & 0.281                     & 0.173                     & 0.314            & 0.264                       & 0.293                                                   & \textbf{0.338}                                                                        \\
wheel    & \textbf{0.683}            & 0.369                     & 0.513                     & 0.246                       & 0.440                                                   & 0.329                                                                        \\ \midrule
average  & 0.386                     & 0.365                     & \textbf{0.415}            & 0.305                       & 0.371                                                   & 0.390                                                                        \\ \bottomrule
\end{tabular}
}
\label{tab:model_variants}
\end{table}

\subsection{Model Selection}
\label{sec:fusion}
\begin{figure}
    \centering
    \includegraphics[width=\textwidth]{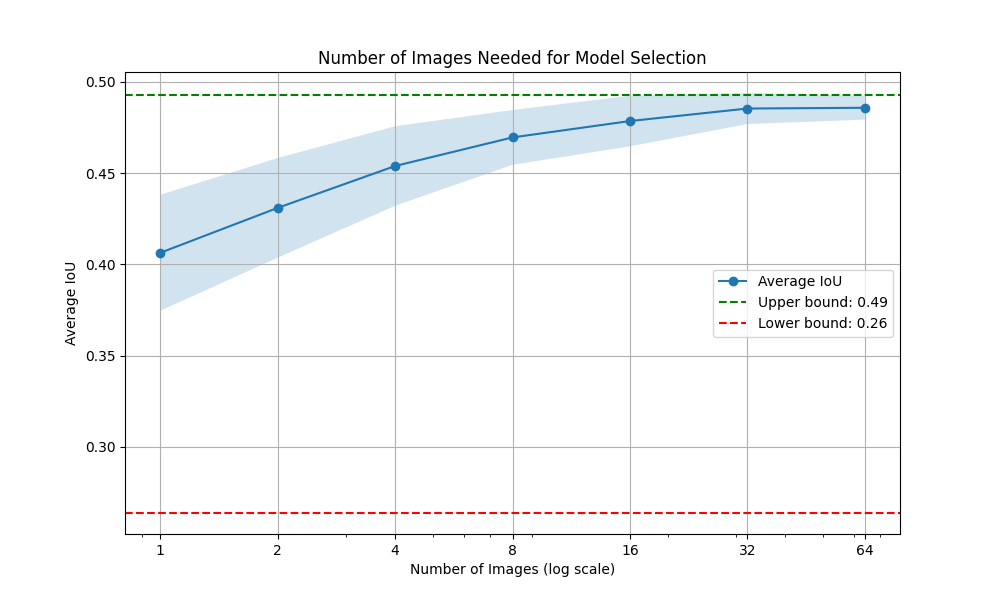}
    \caption{After having seen a few images, e.g. 4 or 8, it is already possible to determine an effective selection of models to acquire better performance by fusion.}
    \label{fig:method-selection}
\end{figure}

There is no single model variant that performs best for all parts (see Table \ref{tab:model_variants}). Therefore, we explore model fusion. When selecting the best scores per part out of the three ROI-based methods we get an IoU of 0.493, which is a great improvement over the best model variant (LGGAM with 0.415). We want to understand how many images are needed to decide properly about this model selection. The upper bound is the best-case scenario, established from having seen the full set. The lower bound is worst-case model selection for each part. Now, we are interested in the performance of model fusion when selecting a model variant for each part, after seeing 1, 2, 4, ..., 64 random images. For each amount of images, the experiment is repeated 10 times, because it involves random draws of the images. The increasing performance is shown in Figure \ref{fig:method-selection}. We observe that the average IoU starts way above the lower bound, indicating that just one image is already an indication of which model is most suitable for respective parts. After having seen a few images, e.g. 4 or 8, it is already possible to determine an effective selection of models to acquire better performance by fusion. With 32-64 images, the performance is close to the upper bound. 

\begin{figure}[h!]
    \centering
    \includegraphics[width=\textwidth]{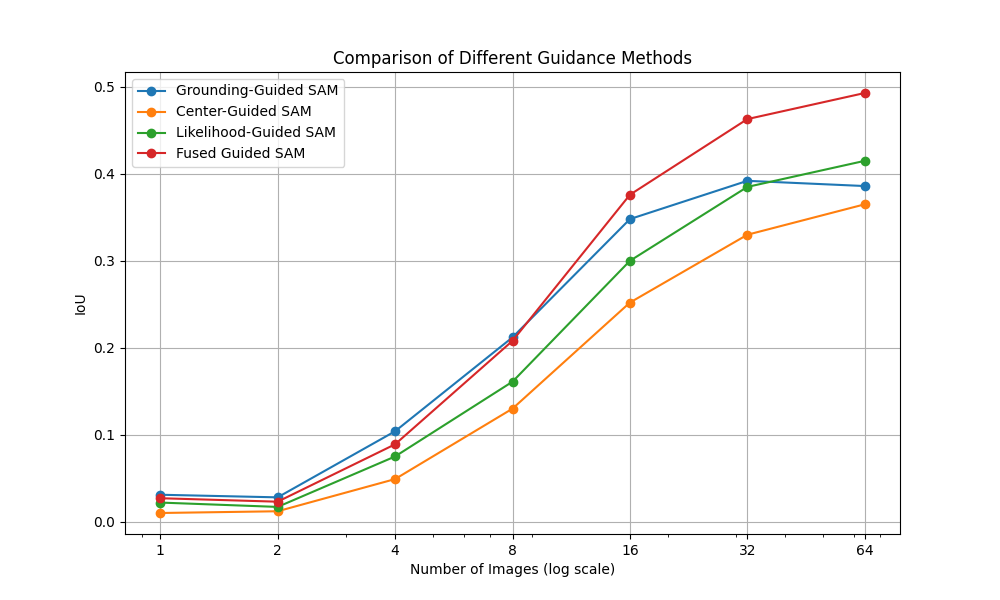}
    \caption{Learning efficiency of our model variants and the fused model. With 64 images, a performance of IoU$\approx$0.49 is achieved. With 32 or only 16 images, performance drops with respectively only 0.04 or 0.12.}
    \label{fig:label-efficiency}
\end{figure}

\subsection{Label Efficiency}

We hypothesize that the performance of Guided SAM largely depends on the accuracy of the guidance classifier. This classifier is trained with 1, 2, 4, ..., 64 images. We explore how many training images are needed for effective guidance. At various amounts of training images, we evaluate the model variants (Section \ref{sec:model_variants_eval}) and the fused model (Section \ref{sec:fusion}), which combines the best performing model variants per part. Figure \ref{fig:label-efficiency} shows the learning efficiency. The fused model is the best performer on average. Our guidance with Grounded SAM, i.e. Grounded Guided SAM (GGSAM), is the best performer at very low number of images. Probably this is because the confidences of that model are a useful source to filter out wrong segmentations. With more than 8 training images, the fused model has a better performance, especially with 16, 32 or 64 images. With more training images, the guidance becomes better, hence all model variants become better. As a consequence, merit can be taken that the best variant is different across parts (Table \ref{tab:model_variants}). With 64 images, a performance of IoU$\approx$0.49 is achieved. With 32 or only 16 images, object parts can be segmented reasonably well: performance drops with respectively only 0.04 or 0.12.

\section{Conclusion}
\label{sec:conclusion}
In this paper, we proposed a novel method for guiding segmentation models to accurately identify object parts. Our approach leverages regions of interest (ROIs) composed of patches predicted by a learnt classifier to identify specific parts of the object and indicate a positional prompt as starting point for part segmentation. It can be used as a guidance for advanced segmentation models such as (Grounded) SAM. We evaluated our method using the Car Parts dataset and demonstrated that it achieves good performance, even with a limited number of labeled patches. This approach significantly reduces the manual effort required for annotation, as it relies on labeling patches rather than creating full segmentation masks. The patch annotations must be centered around the object parts to ensure that the SAM positional prompts are correctly placed. Misalignment could lead the model to segment the background instead of the intended object parts. For future work, we plan to explore techniques to automatically refine patch placement to enhance segmentation accuracy further. Additionally, we aim to extend our method to other datasets and object categories to validate its generalizability and robustness across various domains. 

%
%
\bibliographystyle{splncs04}
\bibliography{main}

\begin{thebibliography}{10}
\providecommand{\url}[1]{\texttt{#1}}
\providecommand{\urlprefix}{URL }
\providecommand{\doi}[1]{https://doi.org/#1}

\bibitem{biederman1987recognition}
Biederman, I.: Recognition-by-components: a theory of human image understanding. Psychological review  \textbf{94}(2), ~115 (1987)

\bibitem{cortes1995support}
Cortes, C., Vapnik, V.: Support-vector networks. Machine learning  \textbf{20},  273--297 (1995)

\bibitem{gupta2019lvis}
Gupta, A., Dollar, P., Girshick, R.: Lvis: A dataset for large vocabulary instance segmentation. In: Proceedings of the IEEE/CVF conference on computer vision and pattern recognition. pp. 5356--5364 (2019)

\bibitem{jain1988evidence}
Jain, A.K., Hoffman, R.: Evidence-based recognition of 3-d objects. IEEE Transactions on Pattern Analysis and Machine Intelligence  \textbf{10}(6),  783--802 (1988)

\bibitem{jia2020fashionpedia}
Jia, M., Shi, M., Sirotenko, M., Cui, Y., Cardie, C., Hariharan, B., Adam, H., Belongie, S.: Fashionpedia: Ontology, segmentation, and an attribute localization dataset. In: Computer Vision--ECCV 2020: 16th European Conference, Glasgow, UK, August 23--28, 2020, Proceedings, Part I 16. pp. 316--332. Springer (2020)

\bibitem{kirillov2023segment}
Kirillov, A., Mintun, E., Ravi, N., Mao, H., Rolland, C., Gustafson, L., Xiao, T., Whitehead, S., Berg, A.C., Lo, W.Y., et~al.: Segment anything. In: Proceedings of the IEEE/CVF International Conference on Computer Vision. pp. 4015--4026 (2023)

\bibitem{liu2023grounding}
Liu, S., Zeng, Z., Ren, T., Li, F., Zhang, H., Yang, J., Li, C., Yang, J., Su, H., Zhu, J., et~al.: Grounding dino: Marrying dino with grounded pre-training for open-set object detection. arXiv preprint arXiv:2303.05499  (2023)

\bibitem{myers2015affordance}
Myers, A., Teo, C.L., Ferm{\"u}ller, C., Aloimonos, Y.: Affordance detection of tool parts from geometric features. In: 2015 IEEE International Conference on Robotics and Automation (ICRA). pp. 1374--1381. IEEE (2015)

\bibitem{nauta2023pip}
Nauta, M., Schl{\"o}tterer, J., van Keulen, M., Seifert, C.: Pip-net: Patch-based intuitive prototypes for interpretable image classification. In: Proceedings of the IEEE/CVF Conference on Computer Vision and Pattern Recognition. pp. 2744--2753 (2023)

\bibitem{oquab2023dinov2}
Oquab, M., Darcet, T., Moutakanni, T., Vo, H., Szafraniec, M., Khalidov, V., Fernandez, P., Haziza, D., Massa, F., El-Nouby, A., et~al.: Dinov2: Learning robust visual features without supervision. arXiv preprint arXiv:2304.07193  (2023)

\bibitem{palmer1999vision}
Palmer, S.E.: Vision science: Photons to phenomenology. MIT press (1999)

\bibitem{DSMLR_Carparts}
Pasupa, K., Kittiworapanya, P., Hongngern, N., Woraratpanya, K.: {Evaluation of deep learning algorithms for semantic segmentation of car parts}. Complex {\&} Intelligent Systems pp. 1--13 (May 2021). \doi{10.1007/s40747-021-00397-8}

\bibitem{radford2021learning}
Radford, A., Kim, J.W., Hallacy, C., Ramesh, A., Goh, G., Agarwal, S., Sastry, G., Askell, A., Mishkin, P., Clark, J., et~al.: Learning transferable visual models from natural language supervision. In: International conference on machine learning. pp. 8748--8763. PMLR (2021)

\bibitem{ramanathan2023paco}
Ramanathan, V., Kalia, A., Petrovic, V., Wen, Y., Zheng, B., Guo, B., Wang, R., Marquez, A., Kovvuri, R., Kadian, A., et~al.: {PACO}: Parts and attributes of common objects. In: Proceedings of the IEEE/CVF Conference on Computer Vision and Pattern Recognition. pp. 7141--7151 (2023)

\bibitem{reddy2018carfusion}
Reddy, N.D., Vo, M., Narasimhan, S.G.: Carfusion: Combining point tracking and part detection for dynamic 3d reconstruction of vehicles. In: Proceedings of the IEEE conference on computer vision and pattern recognition. pp. 1906--1915 (2018)

\bibitem{ren2024grounded}
Ren, T., Liu, S., Zeng, A., Lin, J., Li, K., Cao, H., Chen, J., Huang, X., Chen, Y., Yan, F., et~al.: Grounded sam: Assembling open-world models for diverse visual tasks. arXiv preprint arXiv:2401.14159  (2024)

\bibitem{sun2023going}
Sun, P., Chen, S., Zhu, C., Xiao, F., Luo, P., Xie, S., Yan, Z.: Going denser with open-vocabulary part segmentation. In: Proceedings of the IEEE/CVF International Conference on Computer Vision. pp. 15453--15465 (2023)

\bibitem{wei2024ov}
Wei, M., Yue, X., Zhang, W., Kong, S., Liu, X., Pang, J.: {OV-PARTS}: Towards open-vocabulary part segmentation. Advances in Neural Information Processing Systems  \textbf{36} (2024)

\bibitem{zhao2023fast}
Zhao, X., Ding, W., An, Y., Du, Y., Yu, T., Li, M., Tang, M., Wang, J.: Fast segment anything (2023)

\end{thebibliography}

\end{document}